\documentclass{article}
\usepackage{spconf,graphicx}

\usepackage[utf8]{inputenc}
\usepackage[english]{babel}

\usepackage{amsmath}
\usepackage{amssymb}

\usepackage{dsfont}

\usepackage{xcolor}

\usepackage{booktabs}
\usepackage{tablefootnote}
\usepackage{threeparttable}

\usepackage{pgfplots}
\pgfplotsset{compat = newest}
\usepackage{subcaption}

\usepackage{blindtext}

\newlength{\twosubht}
\newsavebox{\twosubbox}

\title{Text Augmentation for Language Models in High Error Recognition Scenario}
\name{Karel Beneš, Lukáš Burget\thanks{The work was supported by Czech National Science Foundation (GACR) project "NEUREM3" No. 19-26934X.}}
\address{Brno University of Technology}
\begin{document}
\maketitle
\begin{abstract}
We examine the effect of data augmentation for training of language models for speech recognition.
We compare augmentation based on global error statistics with one based on per-word unigram statistics of ASR errors and observe that it is better to only pay attention the global substitution, deletion and insertion rates.
This simple scheme also performs consistently better than label smoothing and its sampled variants.
Additionally, we investigate into the behavior of perplexity estimated on augmented data, but conclude that it gives no better prediction of the final error rate.
Our best augmentation scheme increases the absolute WER improvement from second-pass rescoring from 1.1\,\% to 1.9\,\% absolute on the CHiMe-6 challenge.
\end{abstract}
\section{Introduction}
The traditional reason language models (LMs) appear in ASR systems is that they directly represent the prior term $P(S)$ in the Bayes factorization of the posterior probability $P(S|A)$ of a sentence $S$ given the audio $A$.
However in practice, LMs trained on excessive amounts of data are combined with hybrid and end-to-end systems alike~\cite{Irie2019Transformer,Toshniwal2018EndToEndLM,Cho2019ESPNetLM} at authors liberty.
Overall, LMs can be seen as a refinement tool to apply on a preliminary result of recognition.

\begin{figure*}
    \sbox\twosubbox{%
        \resizebox{\dimexpr.9\textwidth-1em}{!}{%
            \includegraphics[height=3cm]{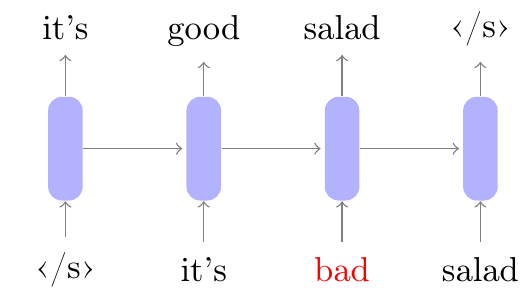}%
            \includegraphics[height=3cm]{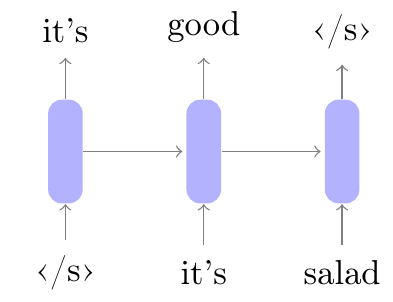}%
            \includegraphics[height=3cm]{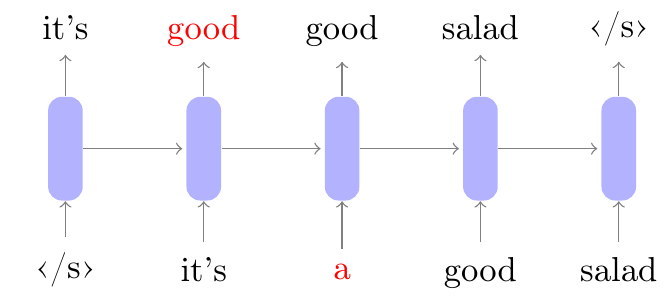}%
        }%
    }
    \setlength{\twosubht}{\ht\twosubbox}
    \centering
    \subcaptionbox{%
        Substitution\label{fig:substitution}}
        {\includegraphics[height=\twosubht]{substitution.pdf}}
    \qquad
    \subcaptionbox{%
        Deletion\label{fig:deletion}}
        {\includegraphics[height=\twosubht]{deletion.pdf}}
    \qquad
    \subcaptionbox{Insertion\label{fig:insertion}}{%
        \includegraphics[height=\twosubht]{insertion.pdf}
    }
    \caption{%
        How errors are introduced for NN LMs.
        The original sentence is ``\emph{it's good salad}''.
        With substitution~(\subref{fig:substitution}), only the input token is replaced.
        To simulate deletions~(\subref{fig:deletion}), both the input and the target at the given position are removed.
        Finally, inserted words~(\subref{fig:insertion}) get into the input while the original target token gets duplicated.
        }\label{fig:error-schemes}
\end{figure*}

To this end, language models can be trained to cope with errors introduced during the first phase of ASR\@.
This is especially well pronounced in discriminative LMs~\cite{RoarkDiscriminative,ZhouDiscriminative,ObaDiscriminative}, which focus on obtaining the final hypothesis from a pool of first-pass hypotheses, oftentimes explicitly taking acoustic clues into account.
On the other hand, we have recently achieved interesting improvements by simply augmenting the training data for a conventional generative LM~\cite{ZmolikovaChime6}.

The idea of working with data similar to ASR output is not new:
Errors are typically introduced in the form of substitutions, driven by a custom \emph{confusability} measure~\cite{SimonnetSimulating}.
Traditionally, this confusability is based on phonemic confusions~\cite{JyothiSimulated},
Recently, sequence-to-sequence models have been proposed for the task~\cite{SeraiSimulation}.

In this work, we elaborate on our central idea that an LM should be capable of good predictions of the next word even when it is exposed to some mistakes in the history:
We extend the idea of augmenting data from substitutions only to deletions as well as insertions.
Then, we investigate the source of improvement by comparing this well-motivated input augmentation to target augmentation.
Finally, we examine the change of the perplexity implied by the data augmentation.

\section{Simulating the Errors}
In the traditional setting, language models are trained to maximize the probability of the word $w_t$ at any position $t$ in the text, as conditioned on the history $h_t$ comprising all previous words $w_1\ldots{}w_{t-1}$:
\begin{equation}\label{eq:ppl}
	- \log \mathrm{PPL} = \frac{1}{T} \sum_{t=1}^{T} \log p(w_t|h_t)
\end{equation}

In this study, we expose the LM to erroneous $h_t$, similar to what it experiences when processing output from an ASR system, giving rise to \emph{simulated PPL}:
\begin{equation}\label{eq:sppl}
	- \log \mathrm{sPPL} = \frac{1}{T} \sum_{t=1}^{T} \mathds{E}_{\hat{h}_t\thicksim{}p_{\mathrm{ASR}}(h_t)} \Big[ \log p(w_t|\hat{h}_t) \Big]
\end{equation}

We discuss the approximations of $p_{\mathrm{ASR}}(h_t)$ in Section~\ref{sec:simulation}.
In general, the input history $h_t$ is processed token by token and individual edits are introduced as illustrated in Fig.~\ref{fig:error-schemes}.
We take care not to remove, replace or introduce sentence boundaries\footnote{Due to a bug in the implementation, we provided no special care to sentence breaks initially. Despite being theoretically unsound, it did not have any observable impact.}.

We also introduce target augmentation to check that the LM benefits from modeling of ASR errors rather than simply from regularization by adding noise to the data. 
Target augmentation differs in that when introducing substitutions, we keep the input token and replace the target one.
Contrasting this to the simulated perplexity (omitting deletions and insertions), we arrive at:

\begin{equation}\label{eq:tppl}
	- \log \mathrm{tPPL} = \frac{1}{T} \sum_{t=1}^{T} \mathds{E}_{\hat{w}} \Big[ \log p(\hat{w}_t|h_t) \Big]
\end{equation}

\subsection{Error simulating distribution}\label{sec:simulation}
As a baseline for error simulation, we do the sampling in a truly flat manner:
We simply roll an unfair 4-sided dice to determine which of the four actions to take.
In case substitution or insertion should be done, we take a sample from uniform distribution over the vocabulary to obtain the new input token.
We call this the \emph{0-gram} error model.
By adjusting the initial categorical distribution (the dice), we have a fine control over the strength of the data augmentation.

In order to better match the actual errors made by the first-pass recognizer, we propose a stronger, \emph{1-gram}, model.
We prepare it as follows:
For a given set of utterances, we get the 100-best hypotheses from the ASR system.
Secondly, we align these hypotheses to the actual transcriptions of this data.
Then for each reference word~$w_{act}$, we summarize its alignments over the whole data and normalize the counts of hypothesised words to get the distribution $p_{sub}(w|w_{act})$.
This categorical distribution is then used to decide the action to apply on every training token.
Note that by treating the empty symbol $\varepsilon$ as a regular word, we also naturally model insertions and deletions this way.
In the case of 1-gram error models, the overall rate of substitutions, deletions and insertions is given by the statistics themselves.

\section{Experiments}\label{sec:experiments}
We evaluate the proposed techniques on Track 1 of the CHiMe-6 challenge~\cite{watanabe_chime6}.
The size of training and development data, including sentence boundaries, is 522k and 136k tokens respectively.
For ease of implementation, we stay with the official large vocabulary of 127k words, letting the output softmax layer to learn that many of those do not occur in the training data.

As the ASR to provide inputs for this paper, we used a single Kaldi system based on a mix of CNN and TDNN-f layers.
The first-pass decoding network is based on a KN-smoothed 3-gram LM\@.
For further details on the design of the ASR system, refer to the system description~\cite{ZmolikovaChime6}.

This system achieves 48.39 \% WER on the development data.
The error is composed respectively of 5\,\%, 17\,\% and 26\,\% of insertions, deletions and substitutions.

For any LM evaluated, we extract 3000-best hypotheses to rescore and use the development set to tune the linear interpolation coefficient for mixing the LSTM LM with the first-pass 3-gram LM\@.
When rescoring, we carry the hidden states over, except for session breaks.
This way, we effectively model the language across segments~\cite{Irie2019AcrossSegment}.

\subsection{Language Model Training}
In all our experiments, we a use two layer LSTM~\cite{Sundermeyer2012LSTM} with 650 units per layer and the dimensionality of input word embeddings reduced to 100.
We train our language models in BrnoLM\footnote{https://github.com/BUTSpeechFIT/BrnoLM}.

We train the LMs with plain SGD, in two stages:
At first, we train the LM from scratch with shuffled lines.
In this stage, we always employ the data augmentation technique under test.
Secondly, we take the trained LM and finetune it on the sentences in their original order.
We ran this stage twice, with this augmentation turned either on or off.
It was always slightly better to do this finetuning with clean data, thus we only report these results.

We begin the first phase training with learning rate 2.0 and start the finetuning with 0.2, in both stages halving the learning rate when development perplexity does not improve.
With target augmentation, we observed the training to be more noisy, therefore we only halved the learning rate when no improvement was observed for 3 consecutive epochs.

\subsection{Tested Augmentation Schemes}
In total, we test LMs trained with seven augmentation schemes:
\begin{enumerate}
    \item The \emph{baseline}, which is only trained on the actual training transcripts.
    \item the 0-gram model (i0), which is trained with uniformly sampled errors. 
    With this model, we optimize for the best rate of substitutions, deletions and insertions.
    \item the 1-gram model (i1), where we collect the statistics from the training data.
    \item the oracle 1-gram model (i1o), where the statistics are collected from the development data.
    \item a 0-gram target augmentation (t0S), where we only introduce substitutions, at the rate optimal for input augmenting systems.
    \item a 0-gram target augmentation (t0SDI), where we introduce deletions and insertion in addition to the target substitution.
    \item target label smoothing (t0LS)~\cite{SzegedyLabelSmoothing}.
    Note that unlike the other techniques, label smoothing requires a principally different change of the training procedure.
\end{enumerate}

For all augmentation schemes, we sweep across the rate of dropout in range [0.0, 0.7] to find the optimal level of total regularization.
For the baseline, the best result comes from setting it to 0.7\footnote{In this case, we tried higher values to check it is the optimum.}, those trained with data augmentation were fairly robust to the dropout rate and achieved their best performance in range of 0.3\,--\,0.6.

\subsection{CHiMe-6 Rescoring Results}
\begin{figure}
    \centering
	\includegraphics[width=\linewidth]{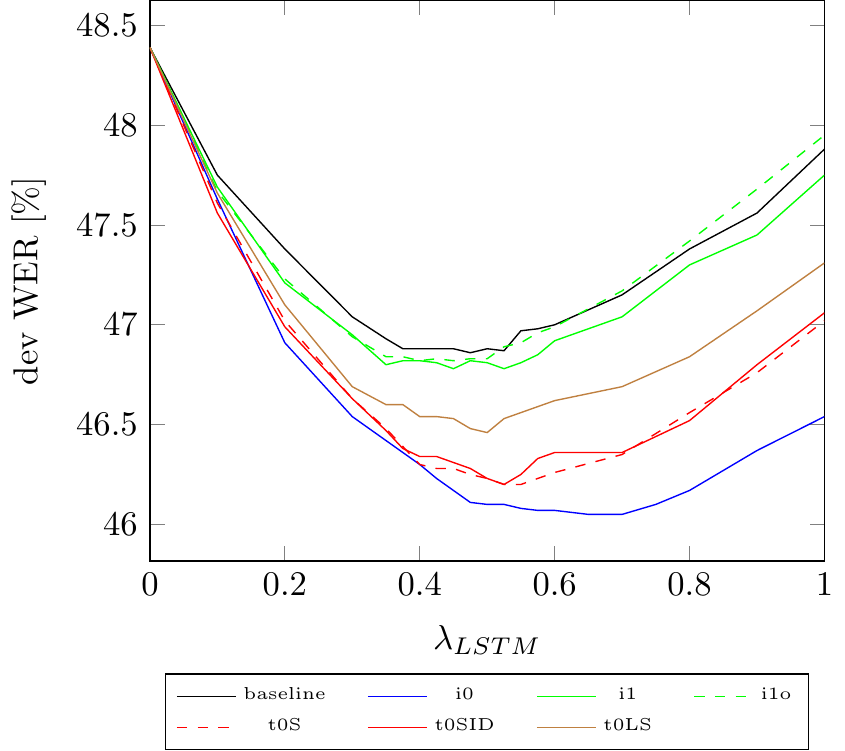}
	\caption{%
	    Development WER as a function of LSTM LM weight.
        For each augmentation scheme, the best dropout rate is selected.
        The left edge represents retaining only the LM score from the original 3-gram LM\@.
        Note how\,--\,with the minor exception of oracle 1-gram input augmentation (i1o)\,--\,the performance of the LSTM LM itself (right edge) corresponds to the performance of its optimal interpolation.
	}\label{fig:results}
\end{figure}

We first assess the performance on the development data, as captured in Figure~\ref{fig:results}.
Overall, we see that the behavior of all LSTM LMs is smooth with respect to the interpolation coefficient.

The best performance is achieved by the input 0-gram augmentation.
For this augmentation, we have found values of around 0.23, 0.15 to work the best as the substitution and deletion rate respectively.
Insertions did not provide any measurable improvements up till 0.1; higher values caused degradation.
This suggests that the input augmentation should roughly correspond to the actual errors rates (see Sec.~\ref{sec:experiments})

On the other hand, neither input 1-gram augmentation was significantly better than the baseline.
The target augmenting LMs do achieve improvement over the baseline, albeit smaller than the 0-gram input augmentation.
Introduction of deletions and insertions brought no improvement for the target 0-gram augmentation, however we have observed these LMs to be rather insensitive to the rate of dropout.

In Table~\ref{tab:results}, we capture the performance of the models on the unseen evaluation data.
We can see that the overall behavior of individual systems stays similar and the input 0-gram augmentation brings clearly the largest gain.

\begin{table}[h]
    \centering
    \caption{%
        Results of rescoring ASR outputs with LMs trained with different data augmentation schemes.
        We report the result of the optimal setting of dropout and LSTM-LM weight as per development results.
    }\label{tab:results}
    \centering
    \begin{tabular}{lrr}
        \toprule
                                    & development       & evaluation \\
        \midrule
        3-gram only                 & 48.39             & 48.82 \\
        baseline                    & 46.86             & 47.69 \\[5pt]
        input 0-gram                & \textbf{46.05}    & \textbf{46.92} \\
        input 1-gram                & 46.85             & 47.70 \\
        input 1-gram oracle         & 46.82             & 47.97 \\[5pt]
        target 0-gram S             & 46.20             & 47.41 \\
        target 0-gram SID           & 46.20             & 47.23 \\
        target label smoothing      & 46.43             & 47.70 \\
        \bottomrule
    \end{tabular}
\end{table}

\begin{figure}
    \centering
	\includegraphics[width=0.75\linewidth]{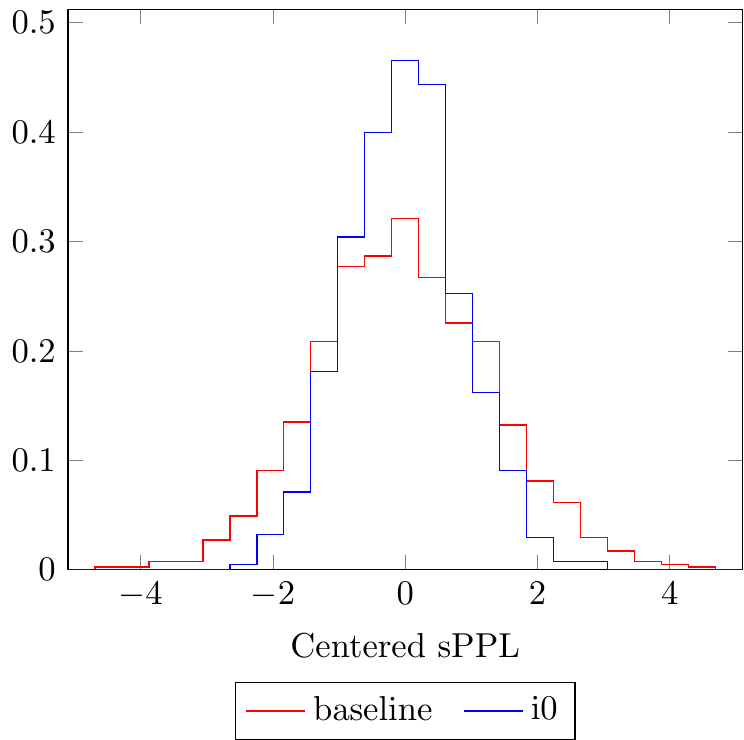}
	\caption{%
		Normalized histogram of sPPL estimates with substitution and deletion rate respectively 0.23 and 0.15.
        Estimated from 1000 runs across the development data.
		Note that values have been centered, baseline LM has mean-sPPL 226, the i0 one trained with matching data augmentation has 178.
	}\label{fig:sppl-stability}
\end{figure}

\begin{figure}
    \centering
	\includegraphics[width=0.85\linewidth]{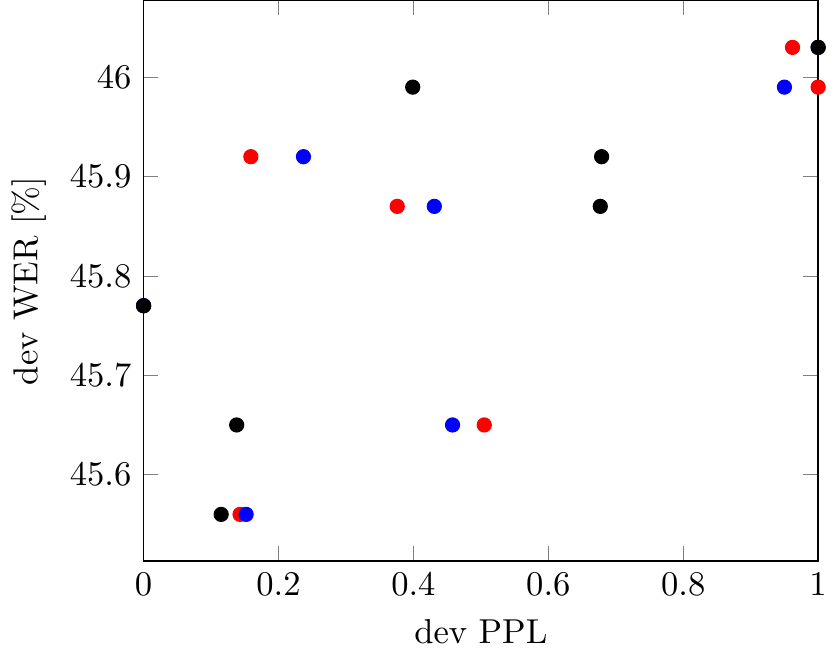}
	\caption{%
		Correlation of different PPL estimates with development WER\@.
        Individual points represent LMs trained with different rate of i0 augmentation.
        Red, blue and black denote respectively sPPL following development error rates, sPPL following training error rates and the vanilla PPL\@.
        For each of the PPL variants, the measured PPL values have been min-max normalized.
	}\label{fig:ppl-vs-wer}
\end{figure}

\subsection{Experiments on Firefighter Speech Recognition}
During the work on this paper, we have applied the 0-gram input augmentation to speech recognition in the OpenSAT challenge~\cite{Karafiat2020opensat}.
This task is considerable easier, with WER at around 10\,\%.
By adding the augmentation, we have achieved a marginal gain of about 0.1\,\% absolute, confirming our hypothesis that this method brings benefit mainly when employed in high error scenarios.

\subsection{Behavior of the Simulated Perplexity}
Since the best performing LMs are trained with the sPPL objective, we investigate two of its properties.
Firstly, we observe its stability as we only estimate the expectation in~\eqref{eq:sppl} from a single realization of noise.
In doing so, we inspect behavior of two different LMs, where one was trained to optimize sPPL and the other regular PPL\@.
Then we examine the predictive power of sPPL towards the final WER of the system.

The stability is captured in Figure~\ref{fig:sppl-stability}.
The sPPLs are approximately normally distributed, with relative standard deviation of around 0.5\,\%.
Comparing the baseline LM to the one trained on the augmented data, we see that the baseline has significantly higher average sPPL (226.5 vs. 178.7) and a slightly higher standard deviation.

To assess the predictive power of sPPL, we plotted a couple of i0 LMs as described by their development (s)PPL and WER in Figure~\ref{fig:ppl-vs-wer}.
We did not find any conclusive evidence that sPPL would serve as a better predictor than PPL\@.

\section{Conclusions}
We have examined several simple text data augmentations for language model training.
Evaluating them by rescoring ASR outputs on the CHiMe-6 challenge, we have achieved the best result when simply introducing uninformed edits into the stream of input tokens.
This improved the WER by 0.8\,\% absolute over rescoring with the baseline LM given by the same neural architecture, but trained on clean text only.
No improvements were achieved with augmentation based on the actual word level confusions produced by the ASR system.
Finally, a control experiment with target augmentations reached approx. 0.5\,\% abs.\ improvement.
From these experimental results, we conclude that while the LMs do benefit from being exposed to ASR-like errors, most of the improvement is coming from training on noised data per se.

We also investigated into the properties of simulated perplexity (sPPL) estimated on the augmented data.
While being rather stable w.r.t.\ the sampling of the word-level noise, we did not see sPPL predict the WER any better than PPL\@.
Furthermore, we always obtained better results when finetuning the LMs to clean data, solidifying our belief that the proposed augmentation scheme should be viewed as a successful regularization technique rather than as an adaptation to a given ASR system.


\section{Acknowledgements}
We thank Martin Kocour for providing the CHiMe-6 ASR system outputs.

\vfill\pagebreak

\bibliographystyle{IEEEbib}
\bibliography{mybib}

\begin{thebibliography}{10}

\bibitem{Irie2019Transformer}
Kazuki Irie, Albert Zeyer, Ralf Schlüter, and Hermann Ney,
\newblock ``{Language Modeling with Deep Transformers},''
\newblock in {\em Proc. Interspeech 2019}, 2019, pp. 3905--3909.

\bibitem{Toshniwal2018EndToEndLM}
S.~{Toshniwal}, A.~{Kannan}, C.~{Chiu}, Y.~{Wu}, T.~N. {Sainath}, and
  K.~{Livescu},
\newblock ``A comparison of techniques for language model integration in
  encoder-decoder speech recognition,''
\newblock in {\em 2018 IEEE Spoken Language Technology Workshop (SLT)}, Dec
  2018, pp. 369--375.

\bibitem{Cho2019ESPNetLM}
J.~{Cho}, S.~{Watanabe}, T.~{Hori}, M.~K. {Baskar}, H.~{Inaguma},
  J.~{Villalba}, and N.~{Dehak},
\newblock ``Language model integration based on memory control for sequence to
  sequence speech recognition,''
\newblock in {\em ICASSP 2019 - 2019 IEEE International Conference on
  Acoustics, Speech and Signal Processing (ICASSP)}, May 2019, pp. 6191--6195.

\bibitem{RoarkDiscriminative}
Brian Roark, Murat Saraclar, and Michael Collins,
\newblock ``Discriminative n-gram language modeling,''
\newblock {\em Computer Speech \& Language}, vol. 21, pp. 373--392, 04 2007.

\bibitem{ZhouDiscriminative}
Zhengyu Zhou, Jianfeng Gao, Frank Soong, and Helen Meng,
\newblock ``A comparative study of discriminative methods for reranking lvcsr
  n-best hypotheses in domain adaptation and generalization,''
\newblock 06 2006, vol.~1, pp. I -- I.

\bibitem{ObaDiscriminative}
Takanobul Oba, Takaaki Hori, Atsushi Nakamura, and Akinori Ito,
\newblock ``Round-robin duel discriminative language models,''
\newblock {\em IEEE Transactions on Audio, Speech \& Language Processing -
  TASLP}, vol. 20, pp. 1244--1255, 05 2012.

\bibitem{ZmolikovaChime6}
Kate\v{r}ina \v{Z}mol\'{i}kov\'{a}, Martin Kocour, Nicol\'{a}s~Federico
  Landini, Karel Bene\v{s}, Martin Karafi\'{a}t, K.~Hari Vydana,
  Alicia~D\'{i}ez Lozano, Old\v{r}ich Plchot, K.~Murali Baskar, J\'{a}n
  \v{S}vec, Ladislav Mo\v{s}ner, Vladim\'{i}r Malenovsk\'{y}, Luk\'{a}\v{s}
  Burget, Bolaji Yusuf, Ond\v{r}ej Novotn\'{y}, Franti\v{s}ek Gr\'{e}zl, Igor
  Sz\H{o}ke, and Jan \v{C}ernock\'{y},
\newblock ``But system for chime-6 challenge,''
\newblock in {\em Proceedings of CHiME 2020 Virtual Workshop}. 2020, pp. 1--3,
  University of Sheffield.

\bibitem{SimonnetSimulating}
Edwin Simonnet, Sahar Ghannay, Nathalie Camelin, and Yannick Est{\`e}ve,
\newblock ``Simulating {ASR} errors for training {SLU} systems,''
\newblock in {\em Proceedings of the Eleventh International Conference on
  Language Resources and Evaluation ({LREC} 2018)}, Miyazaki, Japan, May 2018,
  European Language Resources Association (ELRA).

\bibitem{JyothiSimulated}
Preethi Jyothi and Eric Fosler-Lussier,
\newblock ``Discriminative language modeling using simulated asr errors,''
\newblock 01 2010, pp. 1049--1052.

\bibitem{SeraiSimulation}
P.~{Serai}, P.~{Wang}, and E.~{Fosler-Lussier},
\newblock ``Improving speech recognition error prediction for modern and
  off-the-shelf speech recognizers,''
\newblock in {\em ICASSP 2019 - 2019 IEEE International Conference on
  Acoustics, Speech and Signal Processing (ICASSP)}, 2019, pp. 7255--7259.

\bibitem{watanabe_chime6}
Shinji Watanabe, Michael Mandel, Jon Barker, Emmanuel Vincent, Ashish Arora,
  Xuankai Chang, Sanjeev Khudanpur, Vimal Manohar, Daniel Povey, Desh Raj,
  David Snyder, Aswin~Shanmugam Subramanian, Jan Trmal, Bar~Ben Yair, Christoph
  Boeddeker, Zhaoheng Ni, Yusuke Fujita, Shota Horiguchi, Naoyuki Kanda, and
  Takuya Yoshioka,
\newblock ``{CHiME-6 Challenge}: Tackling multispeaker speech recognition for
  unsegmented recordings,''
\newblock in {\em 6th {International Workshop on Speech Processing in Everyday
  Environments} {(CHiME 2020)}}, 2020.

\bibitem{Irie2019AcrossSegment}
Kazuki Irie, Albert Zeyer, Ralf Schl{\"u}ter, and Hermann Ney,
\newblock ``Training language models for long-span cross-sentence evaluation,''
\newblock in {\em 2019 IEEE Automatic Speech Recognition and Understanding
  Workshop (ASRU)}. IEEE, 2019, pp. 419--426.

\bibitem{Sundermeyer2012LSTM}
Martin Sundermeyer, R.~Schl{\"u}ter, and H.~Ney,
\newblock ``Lstm neural networks for language modeling,''
\newblock in {\em INTERSPEECH}, 2012.

\bibitem{SzegedyLabelSmoothing}
C.~{Szegedy}, V.~{Vanhoucke}, S.~{Ioffe}, J.~{Shlens}, and Z.~{Wojna},
\newblock ``Rethinking the inception architecture for computer vision,''
\newblock in {\em 2016 IEEE Conference on Computer Vision and Pattern
  Recognition (CVPR)}, 2016, pp. 2818--2826.

\bibitem{Karafiat2020opensat}
Martin Karafiát, Murali~Karthick Baskar, Igor Szöke, Hari~Krishna Vydana,
  Karel Veselý, and Jan~"Honza'' Černocký,
\newblock ``But opensat 2019 speech recognition system,'' 2020.

\end{thebibliography}

\end{document}